%% file: EnCo.tex
\def\year{2024}\relax
\documentclass[letterpaper]{article} 
\usepackage{aaai}  
\usepackage{times}  
\usepackage{helvet}  
\usepackage{courier}  
\usepackage[hyphens]{url}  
\usepackage{graphicx} 
\usepackage{natbib}  
\usepackage{caption} 
\DeclareCaptionStyle{ruled}{labelfont=normalfont,labelsep=colon,strut=off} 
\usepackage{algorithm}
\usepackage{algorithmic}

\usepackage{newfloat}
\usepackage{listings}
\floatstyle{ruled}
\newfloat{listing}{tb}{lst}{}
\floatname{listing}{Listing}

\usepackage{amsmath,amsfonts,amssymb}
\usepackage{bm}

\usepackage{booktabs}
\usepackage{multirow} 

\usepackage{algorithm, algorithmic}
\usepackage{enumitem}

\usepackage{arydshln}

\usepackage{epsfig}
\usepackage{graphicx}

\usepackage{color,xcolor}

\usepackage{hyperref}
\usepackage{url}

\title{Rethinking the Paradigm of Content Constraints in Unpaired \\Image-to-Image Translation}
\author{
	Xiuding Cai\textsuperscript{\rm 1 2},
	Yaoyao Zhu\textsuperscript{\rm 1 2},
	Dong Miao\textsuperscript{\rm 1 2},
	Linjie Fu\textsuperscript{\rm 1 2},
	Yu Yao\textsuperscript{\rm 1 2}\thanks{Corresponding author.}\\
}
\affiliations{
	\textsuperscript{\rm 1} Chengdu Institute of Computer Application, Chinese Academy of Sciences, Chengdu, China\\
	\textsuperscript{\rm 2} University of Chinese Academic Sciences, Beijing, China\\
	\{caixiuding20, zhuyaoyao19, miaodong20, fulinjie19\}@mails.ucas.ac.cn, casitmed2022@163.com
}

\begin{document}

\maketitle


\begin{abstract}
	In an unpaired setting, lacking sufficient content constraints for image-to-image translation (I2I) tasks, GAN-based approaches are usually prone to model collapse. Current solutions can be divided into two categories, reconstruction-based and Siamese network-based. The former requires that the transformed or transforming image can be perfectly converted back to the original image, which is sometimes too strict and limits the generative performance. The latter involves feeding the original and generated images into a feature extractor and then matching their outputs. This is not efficient enough, and a universal feature extractor is not easily available. In this paper, we propose EnCo, a simple but efficient way to maintain the content by constraining the representational similarity in the latent space of patch-level features from the same stage of the \textbf{En}coder and de\textbf{Co}der of the generator. For the similarity function, we use a simple MSE loss instead of contrastive loss, which is currently widely used in I2I tasks. Benefits from the design, EnCo training is extremely efficient, while the features from the encoder produce a more positive effect on the decoding, leading to more satisfying generations. In addition, we rethink the role played by discriminators in sampling patches and propose a discriminative attention-guided (DAG) patch sampling strategy to replace random sampling. DAG is parameter-free and only requires negligible computational overhead, while significantly improving the performance of the model. Extensive experiments on multiple datasets demonstrate the effectiveness and advantages of EnCo, and we achieve multiple state-of-the-art compared to previous methods. Our code is available at \href{https://github.com/XiudingCai/EnCo-pytorch}{https://github.com/XiudingCai/EnCo-pytorch}.
\end{abstract}

\section{Introduction}

Image-to-image translation (I2I) aims to convert images from one domain to another with content preserved as much as possible. I2I tasks have received a lot of attention given their wide range of applications, such as style transfer~\cite{ulyanov2016instance},  semantic segmentation~\cite{yu2017dilated,kirillov2020pointrend}, super resolution~\cite{yuan2018unsupervised}, colorization~\cite{zhang2016colorful}, dehazing~\cite{dong2020fd} and image restoration~\cite{liang2021swinir} \emph{etc}. 

In an unpaired setting, lacking sufficient content constraints for the I2I task, using an adversarial loss~\cite{goodfellow2014generative} alone is often prone to model collapse. To ensure content constraints, current generative adversarial networks (GAN)-based approaches can be broadly classified into two categories. One is reconstruction-based solutions. Typical approaches are CycleGAN~\cite{CycleGAN2017} and UNIT~\cite{liu2017unit}. They propose the cycle consistency or shared-latent space assumption, which requires that the transformed image, or the transforming image, should be able to map back to the original image perfectly. However, these assumptions are sometimes too strict~\cite{park2020cut}. For instance, the city street view is converted into a certain pixel-level annotated label, but reconverting a label to a city street view has yet countless possibilities. Such ill-posed setting limits the performance of reconstruction-based GANs, leading to unsatisfactory generations~\cite{chen2020nicegan}. 

Another solution for content constraints is Siamese networks~\cite{bromley1993signature}. Siamese networks are weight sharing neural networks that accept two or more inputs. They are natural tools for comparing entity differences. For the I2I task, the input image and the generated image are fed to some Siamese networks separately, and the content consistency is ensured by matching the output features. CUT~\cite{park2020cut} re-exploit the encoder of the generator as a feature extractor and propose the PatchNCE loss to maximize the mutual information between the patches of the input and generated images and achieve superior performance over the reconstruction-based methods. Some studies~\cite{mechrez2018contextual, zheng2021spatially} repurpose the pre-trained VGG network~\cite{simonyan2015very} as a feature extractor to constrain the feature correlation between the source and generated images. Given the strength and flexibility of Siamese networks, such content-constrained methods are increasingly widely used. However, Siamese network-based GANs mean that the source and generated images need to be fed into the Siamese networks again separately, which entails additional computational costs for training. In addition, an ideal Siamese network that can measure the differences in images well, is not always available.

\begin{figure*}[h!t]
	\centering
	\includegraphics[width=0.75\linewidth]{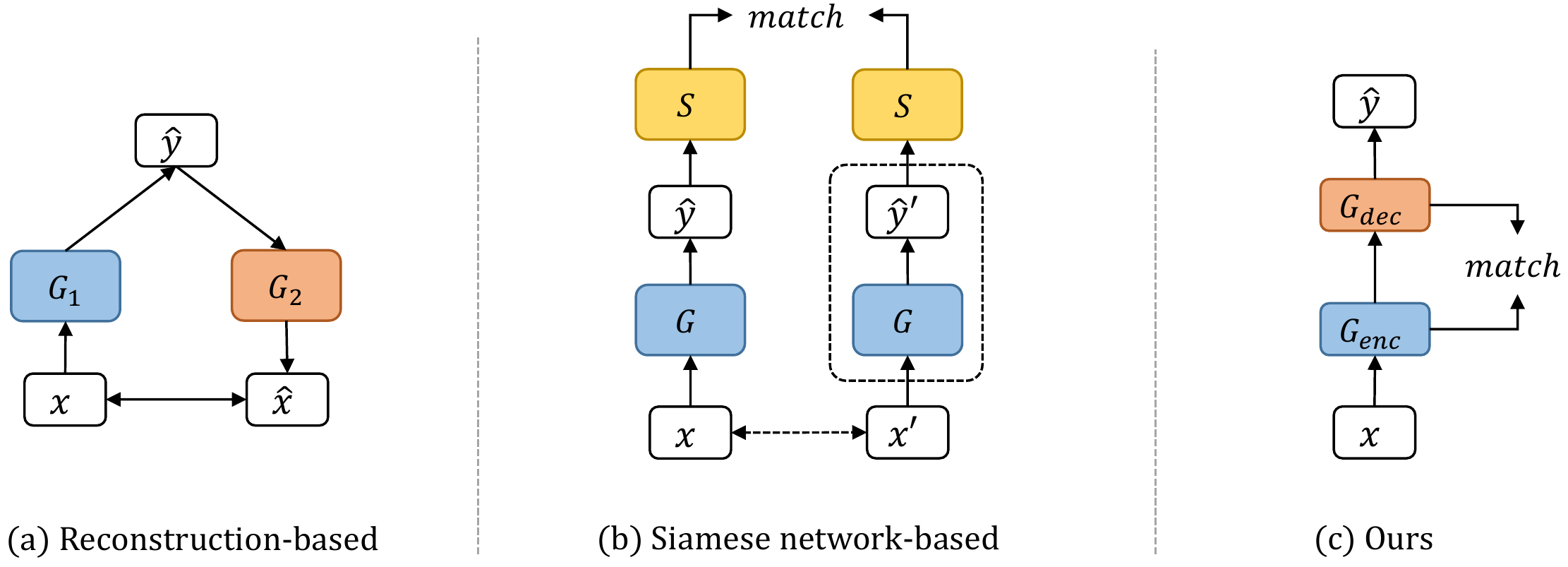}
	\caption{{\bf A comparison of different content constraints frameworks.} (a) Reconstruction-based methods require that $x\leftrightarrow G_2(G_1(x))$, a $\ell_1$ loss or $\ell_2$ loss is always used. Typical methods are CycleGAN~\cite{CycleGAN2017}, UNIT~\cite{CycleGAN2017}, \emph{etc}. (b) Siamese network-based methods like CUT~\cite{park2020cut} or LSeSim~\cite{zheng2021spatially} complete the content constraint through a defined feature extractor $S$, \emph{i.e.}, $\text{match}(S(G(x), x'))$ or $\text{match}(S(G(x), S(G(x')))$, where $x'$ is the augmented $x$. Note that the augmentation of $x$ is optional (dashed box). (c) EnCo completes the content constraint by agreeing on the representational similarity of features from the encoder and decoder of the generator.}
	\label{fig:arch_cmp}
\end{figure*}

Can we explicitly constrain the content inside the generator network? Inspired by U-Net~\cite{ronneberger2015u}, a popular modern network architecture design that integrates features from different stages of the encoder and the decoder by skip connections, we make the encoding-decoding symmetry assumption for the I2I tasks. We assume that the semantic levels of encoder and decoder features from the same stage are the same (note that the number of encoder and decoder stages are opposite).

Based on the assumption, we present EnCo, a simple but efficient way to constrain the content by agreeing on the representational similarity in the latent space of features from the same stage of the \textbf{En}coder and de\textbf{Co}der of the generator. Specifically, we map the multi-stage intermediate features of the network to the latent space through a projection head, in which we aim to bring closer the representation of the same-stage features from the encoder and decoder, respectively. To prevent the networks from falling into a collapse solution, where the projection learns to output constants to minimize the similarity loss, we stop the gradient of the feature branch from the encoder, as well as add a prediction head for the decoder branch. It is worth mentioning that we find that negative samples are not necessary for the EnCo framework for the content constraint. As a result, we use a simple mean squared error loss instead of the contrastive loss that is widely used in current I2I tasks. This eludes the problems associated with negative sample selection~\cite{hu2022qs}.

Benefits from the design, the training of EnCo is efficient and lightweight, as the content constraint is accomplished inside the generative network and no reconstruction or Siamese networks are required. To train more efficiently, similar to CUT, we sample patches from the intermediate features of the generative network, and perform patch-level features matching instead of the entire feature map-level. An ensuing question is from which locations we sample the patches. A simple approach is random sampling, which has been adopted by many methods~\cite{park2020cut, han2021dual, zheng2021spatially}. However, this may not be the most efficient~\cite{hu2022qs}. We note that the discriminator provides key information for the truthfulness of the generated images. However, most of the current GAN-based approaches ignore the potential role of the discriminator in sampling patches (if involved). To this end, we propose a parameter-free discriminative attention-guided patch sampling strategy (DAG). DAG takes advantage of the discriminative information provided by the discriminator and attentively selects more informative patches for optimization. Experimentally, we show that our proposed patch sampling strategy can accelerate the convergence of model training and improve the model generation performance with almost negligible computational effort. 

\begin{itemize}
	\item We propose EnCo, a simple yet efficient way for content constrains by agreeing on the representational similarity of features from the same stage of the encoder and decoder of the generator in the latent space.
	\item We rethink the potential role of discriminators in patch sampling and propose a parameter-free DAG sampling strategy. DAG improves the generative performance significantly while only requiring an almost negligible computational cost.
	\item Extensive experiments on several popular I2I benchmarks reveal the effectiveness and advantages of EnCo. We achieve several state-of-the-art compared to previous methods. 
\end{itemize}


\section{Related Works}

\noindent \textbf{Image-to-Image Translation}
Image-to-image translation ~\cite{isola2017image, wang2018pix2pixHD, CycleGAN2017, park2020cut, wang2021instance} aims to transform images from the source domain to the target domain with the semantic content preserved. Pix2Pix~\cite{isola2017image} was the first framework to accomplish the I2I task using paired data with an adversarial loss~\cite{goodfellow2014generative} and a reconstruction loss. However, paired data across domains are infeasible to be collected in most settings. Methods such as CycleGAN~\cite{CycleGAN2017}, DiscoGAN~\cite{kim2017disco} and DualGAN~\cite{yi2017dualgan} extend the I2I task to an unsupervised setting based on cycle consistency assumption that the generated image should be able to be converted back to the original image again. In addition, UNIT~\cite{liu2017unit} and MUNIT~\cite{huang2018multimodal} propose to learn a shared-latent space in which the hidden variable, \emph{i.e.}, the encoded image, can be decoded as both the target image and the original image. These methods can be classified as reconstruction-based solutions, and they implicitly assume that the process of conversion should be able to reconvert to the original image. However, perfect reconstruction is unlikely to be possible in many cases, which can potentially limit the performance of the generative networks~\cite{park2020cut}. In addition, such methods usually require additional auxiliary generators and discriminators.

\noindent \textbf{Siamese Networks for Content Constraints}
Another solution for content constraints can be attributed to Siamese network-based approaches, and they can effectively address the challenges posed by reconstruction-based ones. Siamese networks usually consist of networks with shared weights that accept two or more inputs, extract features, and compare differences. The selection of Siamese networks for content constraints can be different. For instance, the Siamese network of DistanceGAN and GcGAN is their generator. They require that the distances between the input images and the distances between the output images after generation should be consistent. CUT reuses the encoder of the generator as the Siamese network and proposes PatchNCE loss, aiming to maximize the mutual information between the patches of the input and output images. DCL~\cite{han2021dual} extends CUT to a dual-way settings that exploiting two independent encoders and projectors for input and generated images respectively, but doubles the number of network parameters. Some recent studies have also attempted to re-purposed the pre-trained VGGNet~\cite{simonyan2015very} as a perceptual loss to require that the input and output images should be visually consistent~\cite{zheng2021spatially}. There may be a priori limitations in these methods, such as the frozen network weights of the loss function cannot adapt to the data and thus may not be the most appropriate~\cite{zheng2021spatially}. Our work is quite different from the current approaches, as shown in Figure \ref{fig:arch_cmp}, where we impose constraints inside the generative network, \emph{i.e.}, between the encoder and decoder, without requiring additional networks for reconstruction or feature extraction. Therefore, EnCo has a higher training efficiency.

\noindent \textbf{Contrastive Learning}
Recently, contrastive learning (CL) has achieved impressive results in the field of unsupervised representation learning~\cite{hjelm2018learning, chen2020simple, he2020momentum, henaff2020data, oord2018representation}. Based on the idea of discriminative, CL aims to bring closer the representation of two correlated signals (known as positive pair) in the embedding space while pushing away the representation of uncorrelated signals (known as negative pair). CUT first introduced contrastive learning to the I2I task and has been continuously improved since then~\cite{han2021dual, hu2022qs, zhan2022modulated}. QS-Attn~\cite{hu2022qs} improved the negative sampling strategy of CUT, by computing the $QKV$ matrix to dynamically selects relevant anchor points as positive and negatives. MoNCE~\cite{zhan2022modulated} proposed modulated noise contrastive estimation loss to re-weight the pushing force of negatives adaptively according to their similarity to the anchor. However, the performance of CL-based GANs approaches is still affected by the negative sample selection and poor negative may lead to slow convergence and even counter-optimization~\cite{robinson2020contrastive}. Therefore, some studies have raised the question whether using of the negative is necessary. BYOL~\cite{byol2020} successfully trained a  discriminative network using only positive pairs with a moment encoder. SimSiam~\cite{simsiam2020} pointed out that the stopping gradient is an important component for successful training without negatives, thus removing the moment encoder. EnCo is trained without negatives and only considerate the same stage features from the encoder and decoder of the generator as a positive pair, ensuring content consistency. 

\section{Methods}


\subsection{Main Idea}

Given an image from the source domain ${x}\in\mathcal X$, our goal is to learn a mapping function (also called a generator) $G_{\mathcal{X} {\rightarrow} \mathcal{Y}}$ that converts the image from the source domain to the target domain $\mathcal Y$, \emph{i.e.}, $\hat y=G_{\mathcal X\rightarrow\mathcal Y}(x)$, and with as much content semantic information preserved as possible. 

\begin{figure*}[h!t]
	\centering
	\includegraphics[width=0.98\linewidth]{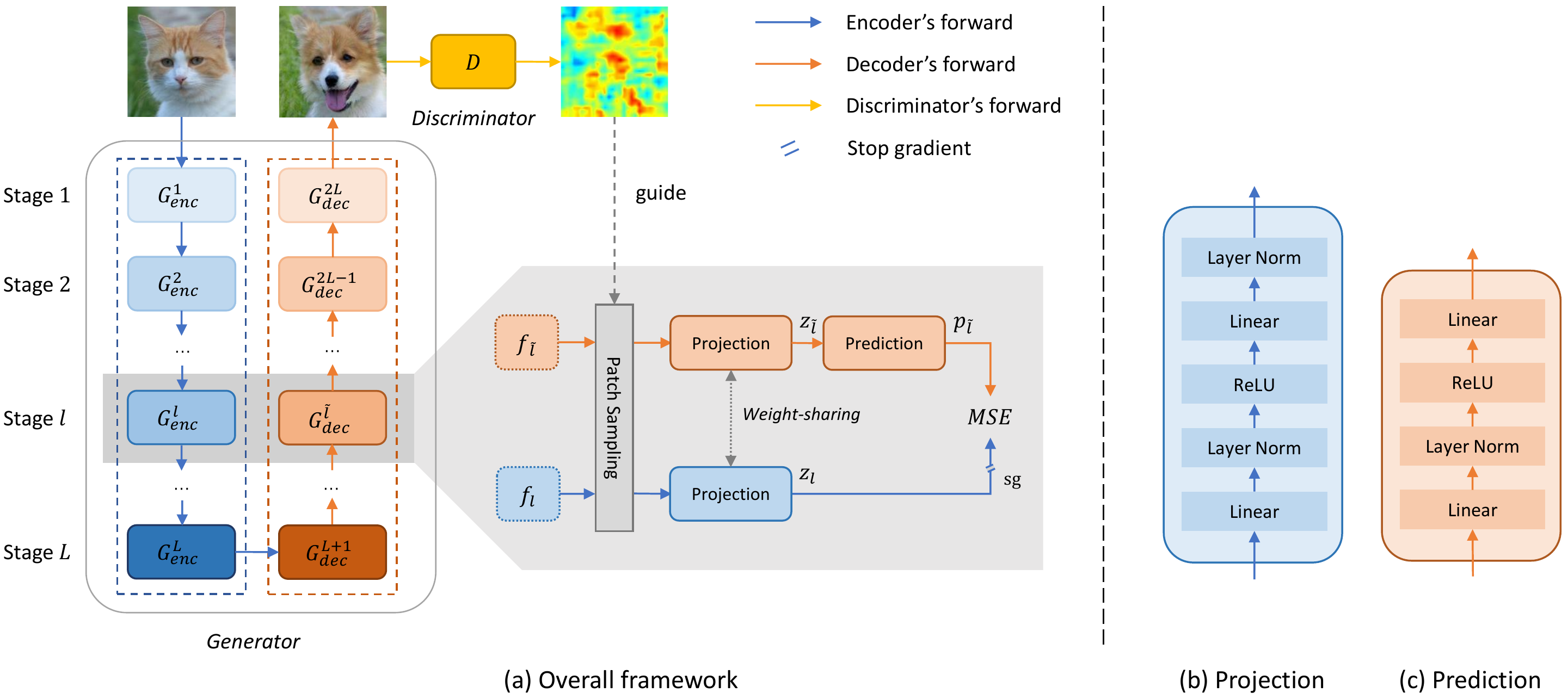}
	\caption{(a) The overview of EnCo framework. EnCo constrain the content by agreeing on the representational similarity in the latent space of features from the same stage of the encoder and decoder of the generator. (b) The architecture of the projection. (c) The architecture of the prediction.
	}
	\label{fig:main_arch}
\end{figure*}

Traditional content constraint methods based on Siamese networks, such as CUT, intend to constrain the content consistency of an image after generation with the source image. EnCo aims to constrain the content consistency of features generated in the intermediate process from the source image to the target image. Our approach is more efficient to train and achieves better performance. The overall architecture is shown in Figure~\ref{fig:main_arch} and contains the generator $G$, the discriminator $D$, the projection head $h$, and the prediction head $g$. We decompose the generator into two parts, the encoder and the decoder, each of which consists of $L$-stage sub-networks. For any input source domain image $x$, after passing through $L$-stage sub-networks of the encoder, a sequence of features of different semantic levels are produced, \emph{i.e.}, $\{h_{l}\}_1^L=\{G_{enc}^l(h_{l-1})\}_1^L$. where $x=h_0$. Then feeding the output of the last stage of the encoder, \emph{i.e.}, $h_{L}$, into the decoder, we can also obtain a sequence of features of different semantic levels in the decoder $\{h_{l}\}_{L+1}^{2L}=\{G_{dec}^l(h_{l-1})\}_{L+1}^{2L}$, where $\hat y=h_{2L}$. For the I2I task, we make the encoding-decoding symmetry assumption that the semantic levels of the encoder and decoder features $f_l$ and $f_{2L-l}$ from the same stage are the same (note that the number of stages of the encoder and decoder is opposite). For brevity, we abbreviate $2L-l$ as $\tilde l$ and denote $(f_l,f_{\tilde l})$ as a pair of same-stage features in the following.

We consider that the content of the transformed image can be preserved by constraining $(f_l,f_{\tilde l})$. However, this direct approach may degrade the optimization of the generative network. With this view, we propose to guarantee the content constraint by constraining the representational similarity of the encoder and decoder features of the generator in the latent space. 

As shown in Figure~\ref{fig:main_arch}, for any pair of same-stage features $(f_l,f_{\tilde l})$, we map them to the $K$-dimensional latent space by a shared two-layer projection head $h(\cdot)$ to obtain $z_l\triangleq h(f_l)$ and $z_{\tilde l}\triangleq h(f_ {\tilde l})$. Inspired by~\cite{byol2020}, we further add a prediction head $g(\cdot)$ to $z_{\tilde l}$ to enhance the non-linear expression of $z_{\tilde l}$, and obtain $p_{\tilde l}\triangleq g(z_{\tilde l})$.

To avoid collapsing or expanding, we $\ell_2$-normalize both $z_l$ and $p_{\tilde l}$ and map them to the unit sphere space to obtain $\bar{z_l}\triangleq z_l/\|z_l\|_2$ and $\bar{p}_{\tilde{l}}\triangleq p_{\tilde l}/\|p_{\tilde l}\|_2$. Finally, we define the following mean-squared error loss aiming to constrain the representational similarity of a pair of same-stage normalized hidden variables from the encoder and decoder,
\begin{equation}
	\mathcal L(z_{\tilde l},z_l) = \|\bar{p}_{\tilde l}-\bar{z_l}\|_2^2
	= 2-2\cdot\frac{\langle g(z_{\tilde l}), z_l\rangle}{\|g(z_{\tilde l})\|_2 \cdot \|z_l\|_2}.
	\label{eq:original_mse}
\end{equation}

To prevent a collapse solution, \emph{i.e.}, the networks learn to output constants to minimize the loss. Following~\cite{simsiam2020}, we solve this problem by introducing the key component of stopping gradient. We modify Eq. \eqref{eq:original_mse} as follows,
\begin{equation}
	\mathcal L(z_{\tilde l},\mathtt{stopgrad}(z_l)).
	\label{eq:sg_mse}
\end{equation}

This means that during optimization, $z_l$ is constant and $z_{\tilde l}$ is expected to be able to predict $z_l$ through the prediction head $g(\cdot)$. Therefore, $z_{\tilde l}$ cannot vary too much from $z_l$, in such a way that the content constraint is achieved.

\subsection{Multi-stage, Patch-based Content Constraints}

Consider a more efficient training approach, given features pair $(f_l,f_{\tilde l})$, we sample $S$ patches from different positions of $f_{\tilde l}$, feed to the projection and prediction head, and obtain the set $\mathbf{q_{\tilde l}}=\{q_{\tilde l}^{(1)},\cdots,q_{\tilde l}^{(S)}\}$, where the subscript indicates which stage to sample and the superscript denotes where to sample from the feature map. Similarly, we can sample from the same position, from $f_l$, and feed to the projection to get $\mathbf{k_l}=\{k_{\tilde l}^{(1)},\cdots,k_{l}^{(S)}\}$. We implement content constraints on patch-level features, rather than the entire feature map-level. Therefore, we just need to use $\mathbf{k_l},\mathbf{q_{\tilde l}}$ to replace $z_l,z_{\tilde l}$ in Eq. \eqref{eq:sg_mse}, respectively, and get
\begin{equation}
	\mathcal L(\mathbf{q_{\tilde l}},\mathtt{stopgrad}(\mathbf{k_l})).
	\label{eq:patch_mse}
\end{equation}

We can further extend Eq. \eqref{eq:patch_mse} to a multi-stage version, \emph{i.e.},
\begin{equation}
	\begin{split}
		&\mathcal{L}_{\mathrm{MultiStage}}(G, h, g, \mathbf{X}) = \\
		&\mathbb{E}_{x \sim \mathbf{X}} \sum_{l}^{\mathbb L} \sum_{s}^{\mathbb{S}_l} \mathcal L({q_{\tilde l}^{(s)}},\mathtt{stopgrad}({k_l^{(s)}})),
	\end{split}
	\label{equ:ncc_loss}
\end{equation}
where, $\mathbb L$ is the set of chosen same-stage pairwise features to calculate the mean-squared error loss, and $\mathbb{S}_l$ is the set of sampled positions of patches from $(f_l,f_{\bar l})$.

\subsection{Discriminative Attention-guided Patch Sampling Strategy}

We further propose an efficient discriminative attention-guided (DAG) patch sampling strategy to replace the current widely used random sampling strategy used in Eq. \eqref{equ:ncc_loss}. The idea of DAG is simple. DAG mainly takes good advantage of the important information from the discriminator: the truthfulness of the generated images, and attentively selects more informative patches for optimization.

Assuming that a total of $K$ patches would be sampled, for any pairwise features $(f_l,f_{\tilde l})$, DAG proceeds as follows: 1) obtaining the attention scores: interpolating the output of the discriminator to the same resolution size as $f_l$ and $f_{\tilde l}$, thus each position on $f_l$ receives a attention score; 2) oversampling: uniformly sampling $kK (k>1)$ patches from $f_{\tilde l}$, where $k$ is the oversampling ratio; 3) ranking: sorting all sampled patches in ascending order according to their corresponding attention scores; 4) importance sampling: selecting the top $\beta K (0\le\beta\le1)$ patches with the highest scores, where $\beta$ is the importance sampling ratio; 5) Covering: uniformly sample the remaining $(1-\beta)K$ patches. Note that the DAG is parameter-free, while requiring only an almost negligible computational cost. 

\begin{figure*}[ht]
	\centering
	\includegraphics[width=0.82\linewidth]{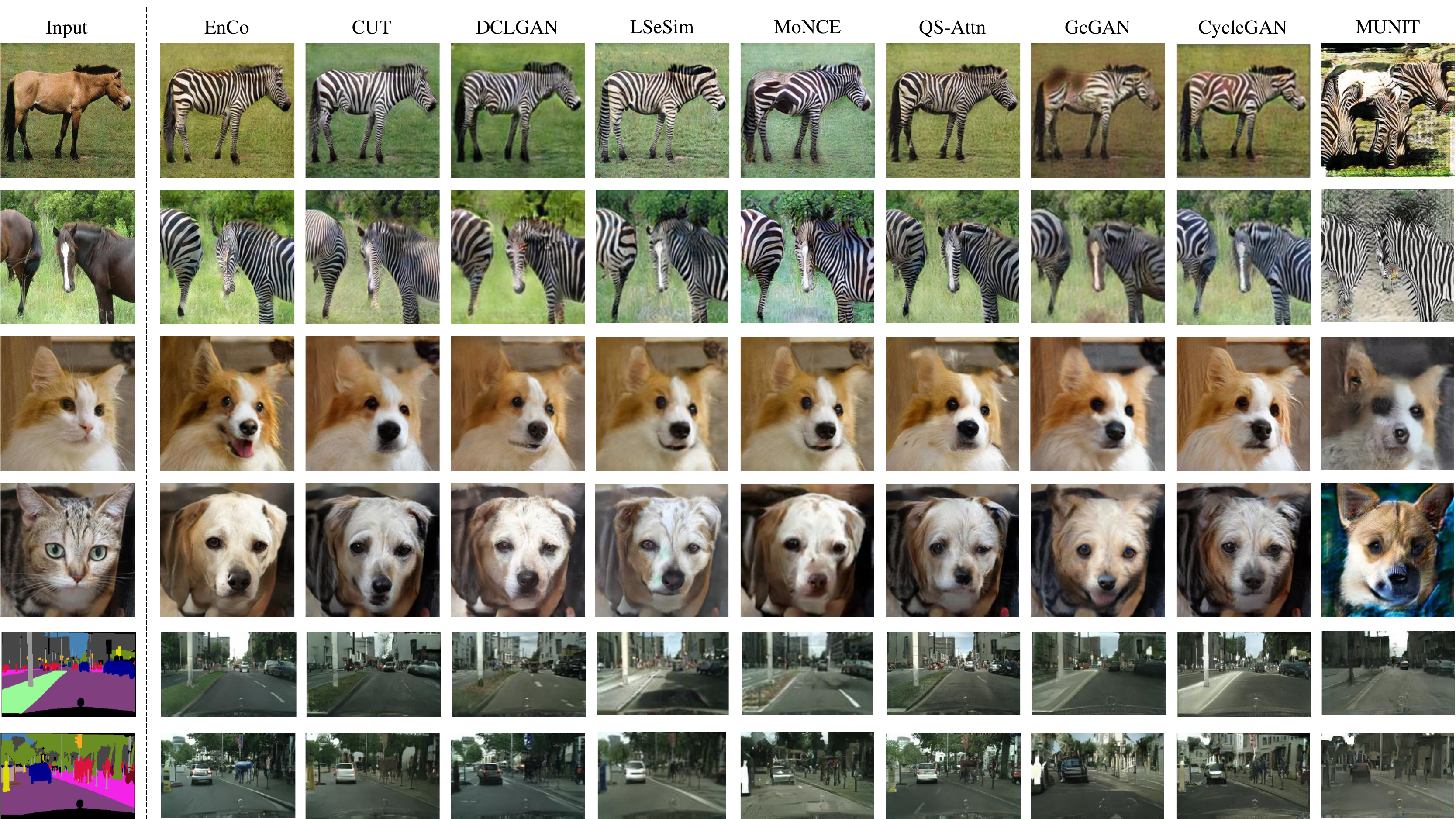}
	\caption{{\bf Results of qualitative comparison.} We compare EnCo with existing methods on the \emph{Horse$\rightarrow$Zebra}, \emph{Cat$\rightarrow$Dog}, and \emph{Cityscapes} datasets. EnCo achieves more satisfactory visual results. For example, in the case of \emph{Cat$\rightarrow$Dog}, EnCo generates a clearer nose for the dog. And in the case of \emph{Cityscapes}, EnCo successfully generates the traffic cone represented in yellow in the semantic annotation, while the other methods yielded only suboptimal results.
	}
	\label{fig:cmp_main}
\end{figure*}

\begin{table*}[h!t]
	\fontsize{10pt}{12pt}\selectfont
	\centering
	\renewcommand{\arraystretch}{1.1}
	\resizebox{0.95\textwidth}{!}{
		\begin{tabular}{{l}*{9}{c}}
			\toprule
			\multirow{2}*{\textbf{Method}} & \multicolumn{4}{c}{\textbf{CityScapes}} & \multicolumn{1}{c}{\textbf{Cat$\rightarrow$Dog}} & \multicolumn{3}{c}{\textbf{Horse$\rightarrow$Zebra}} \\
			\cmidrule(lr){2-5}\cmidrule(lr){6-6}\cmidrule(lr){7-9}
			& \textbf{mAP}$\uparrow$ & \textbf{pixAcc}$\uparrow$ & \textbf{classAcc}$\uparrow$ & \textbf{FID}$\downarrow$ & \textbf{FID}$\downarrow$ & \textbf{FID}$\downarrow$ & \textbf{Mem(GB)}$\downarrow$ & \textbf{sec/iter}$\downarrow$ \\
			\midrule
			CycleGAN~\cite{CycleGAN2017} & 20.4 & 55.9 & 25.4 & 68.6 & 85.9 & 66.8 & 4.81 & 0.40 \\ 
			MUNIT~\cite{huang2018multimodal} & 16.9 & 56.5 & 22.5 & 91.4 & 104.4 & 133.8 & 3.84 & 0.39\\ 
			\cdashline{1-9}
			GCGAN~\cite{geometry2019} & 21.2 & 63.2 & 26.6 & 105.2 & 96.6 & 86.7 & \textbf{2.67} & 0.62\\
			
			CUT~\cite{park2020cut} & 24.7 & 68.8 & 30.7 & 56.4 & 76.2 & 45.5 & 3.33 & 0.24\\
			DCLGAN~\cite{han2021dual} & 22.9 & 77.0 & 29.6 & \underline{49.4} & \underline{60.7} & 43.2 & 7.45 & 0.41\\ 
			FSeSim~\cite{zheng2021spatially} & 22.1 & 69.4 & 27.8 & 54.3 & 87.8 & 43.4 & 2.92 & \underline{0.17}\\ 
			
			MoNCE~\cite{zhan2022modulated} & \underline{25.6} & \underline{78.4} & \underline{33.0} & 54.7 & 74.2 & 41.9 & 4.03 & 0.28  \\ 
			QS-Attn~\cite{hu2022qs} & 25.5 & \textbf{79.9} & 31.2 & 53.5 & 72.8 & \underline{41.1}  & 2.98 & 0.35\\ 
			\cdashline{1-9}
			
			EnCo (Ours) & \textbf{28.4} & {77.3} & \textbf{37.2} & \textbf{45.4} & \textbf{54.7} & \textbf{38.7} & \underline{2.83} & \textbf{0.14} \\ 
			
			\bottomrule
		\end{tabular}
	}
	\caption{{\bf Comparison with baselines on unpaired image translation.} We compare our approach to the state-of-the-art methods on three datasets. We show multiple metrics, where the $\uparrow$ indicates higher is better and the $\downarrow$ indicates lower is better. It is worth noting that our method outperforms all baselines on the FID metric and shows superior results on the \emph{Cityscapes} for the semantic segmentation metric. Also, our method shows a fast training speed.}
	\label{tab:cmp_main}
\end{table*}

\subsection{Full Objective}

In addition to the MultiStage loss presented above, we also use an adversarial loss to complete the domain transfer, and we add an identity mapping loss as a regularization term to stabilize the network training.

\noindent \textbf{Generative adversarial loss}
We use LSGAN loss~\cite{mao2017least} as the adversarial loss to encourage the generated images that are as visually similar to images in the target domain as possible, which is formalized as follows,
\begin{equation}
	\begin{split}
		\mathcal L_{\textrm{GAN}}=\mathbb{E}_{y\sim Y}[D(y)^2]
		+\mathbb{E}_{x\sim X}[(1-D(G(x)))^2].
	\end{split}
	\label{equ:gan_loss}
\end{equation}

\noindent \textbf{Identity mapping loss}
In order to stable the training and accelerate the convergence, we add an identity mapping loss.
\begin{equation}
	\begin{split}
		\mathcal{L}_{\textrm{identity}}(G)=\mathbb E_{y\sim Y}\|G(y)-y\|_1.
	\end{split}
	\label{equ:identity_loss}
\end{equation}
We only use this regular term in the first half of the training phase because we find that it impacts the generative performance of the network to some extent.

\noindent \textbf{Overall loss}
Our final objective function is as follows:
\begin{equation}
	\begin{split}
		\mathcal{L}_\textrm{total}(G,D,h,g) &= \mathcal L_\textrm{GAN}(G,D,X,Y) \\
		&+\lambda_{NCE}\mathcal{L}_{\mathrm{MultiStage}}(G, h, g, {X}) \\
		&+\lambda_{IDT}\mathcal{L}_{\textrm{identity}}(G, Y),
	\end{split}
	\label{equ:total_loss}
\end{equation}
where $\lambda_{NCE}$ and $\lambda_{IDT}$ are set to 2 and 10, respectively.

\subsection{Discussion}

EnCo achieves content consistency by constraining the similarity between representations of the encoder and decoder features at multiple stages in the latent space. In fact, there are two additional perspectives on how EnCo achieves content consistency that we would like to offer. Firstly, in relation to reconstruction-based methods, EnCo requires decoding features that should be able to predict their corresponding encoded features in turn through the prediction MLP, which is somehow similar to the reconstruction-based approach. \textit{However, EnCo conducts the reconstruction at the feature level rather than the pixel level, which provides more freedom when enforcing content consistency.} EnCo degenerates into a special CycleGAN approach when we only constrain the consistency of the input and generated images, with projection being an identity network and prediction network being another generator. Secondly, EnCo can also be regarded as an implicit and lightweight Siamese network paradigm (similar to CUT), where encoder and decoder features are constrained to have similar representations in the latent space through the shared projection MLP.


\section{Experiments}

\subsection{Experiment Setup}

\noindent \textbf{Datasets}
To demonstrate the superiority of our method, we trained and evaluated on three popular I2I benchmark datasets, including \emph{Cityscapes}, \emph{Cat$\rightarrow$Dog}, \emph{Horse$\rightarrow$Zebra}. \emph{Cityscapes}~\cite{cordts2016cityscapes} contains street scenes from German cities with 2975 training images and 500 test images. Each image in \emph{Cityscapes} has a resolution of $2048\times1024$ with high quality pixel-level annotation. \emph{Cat$\rightarrow$Dog} comes from the AFHQ dataset~\cite{choi2020stargan}, containing 5153 images of cats and 4739 images of dog. \emph{Horse$\rightarrow$Zebra}, collected by CycleGAN from ImageNet~\cite{deng2009imagenet}, contains 1067 images of horses and 1334 images of zebras. For all experiments, we resized images to $256\times 256$ resolution size.

\noindent \textbf{Implementation details}
We use the same ResNet-based generator~\cite{park2020cut} and PatchGAN discriminator~\cite{isola2017image} with a receptive field of $70\times70$ as CUT. We use Adam optimizer~\cite{kingma2014adam} with $\beta_1=0.5$ and $\beta_2=0.999$. For the \emph{CityScapes} and \emph{Horse$\rightarrow$Zebra} datasets, 400 epoches are trained, and 200 epoches are trained only for the \emph{Cat$\rightarrow$Dog} dataset. Following TTUR~\cite{heusel2017ttur}, we set unbalanced learning rates of $5e-5$, $2e-4$ and $5e-5$ for the generator, discriminator and projection head, respectively. We start linearly decaying the learning rate halfway through the training with batch size of 1. We use by default a two-layer projection and a two-layer prediction with a dimension of 256 for all linear layers, the same as used in the CUT, but we add layer normalization after the linear layers, except for the last linear layer of the prediction (See Figure~\ref{fig:main_arch} (b) and (c)).

\noindent \textbf{Evaluation metrics}
We mainly use Fr\'{e}chet Inception Distance (FID)~\cite{heusel2017ttur} to evaluate the visual quality of the generated images. FID is one of the most commonly used distribution-based image quality assessment metrics, by comparing the distance between distributions of generated and real images in a deep feature domain. For the \emph{CityScapes} dataset, since they have corresponding segmentation labels, following CUT, we apply a pre-trained segmentation model~\cite{yu2017dilated} to segment the generated images and use three evaluation metrics, including mean average precision (mAP), pixel-wise accuracy (pixAcc), and average class accuracy (classAcc) to measure how well methods discover correspondences. In addition, we also include the training speed and memory consumption to measure the training efficiency of the model.

\subsection{Comparison with the State-of-the-art Methods}

We compare our method with several state-of-the-art methods of unpaired I2I, including CycleGAN~\cite{CycleGAN2017}, MUNIT~\cite{huang2018multimodal}, GcGAN~\cite{geometry2019}, CUT~\cite{park2020cut}, DCLGAN~\cite{han2021dual}, FSeSim~\cite{zheng2021spatially}, MoNCE~\cite{zhan2022modulated} and QS-Attn~\cite{hu2022qs}. Among them, CycleGAN and MUNIT are reconstruction-based methods, and the rest are Siamese network-based ones. 

\noindent \textbf{Quantitative and qualitative results}
The results of the quantitative comparisons are shown in Table~\ref{tab:cmp_main}. As can be seen from Table~\ref{tab:cmp_main}, EnCo outperforms all baselines on FID metric on three datasets. In particular, on tasks \emph{Cityscapes} and \emph{Cat$\rightarrow$Dog}, EnCo's FID drops by 4 and 6 points, respectively, compared to the second place (indicated by underlining). Figure~\ref{fig:cmp_main} gives a qualitative comparison. It can be seen that EnCo achieves more satisfactory generation results. For example, in the case of \emph{Cat$\rightarrow$Dog}, EnCo succeeds in generating a clear dog's nose, but most other methods generate only suboptimal results.

\begin{figure}[ht]
	\centering
	\includegraphics[width=\linewidth]{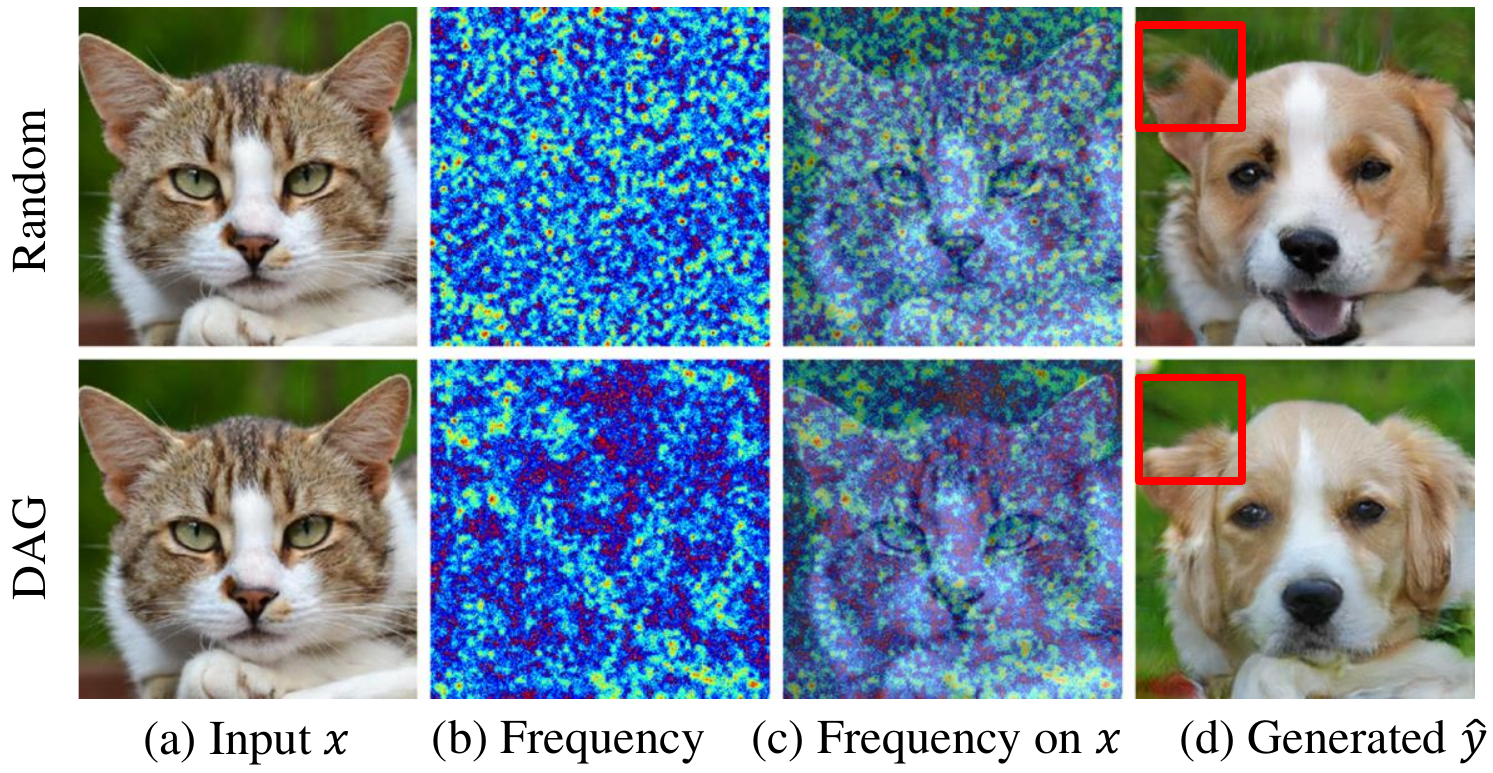}
	\caption{Comparison of different patch sampling strategy. For the input image $x$, we superimpose the sampling positions every 10 epochs to obtain the sampling frequency map (b). As can be seen from (c), compared to the random strategy, our proposed DAG patch sampling strategy are more focused on regions that help in domain discrimination, such as ears, eyes, nose, \emph{etc}. As a result, the model with DAG sampling strategy generates more adorable results than random sampling strategy (see the red box in (d)).
	}
	\label{fig:vis_sampling}
\end{figure}

\begin{table}[h]
	\fontsize{10pt}{12pt}\selectfont
	\centering
	\resizebox{1\columnwidth}{!}{
		\begin{tabular}{{l}*{4}{c}}
			\toprule
			\multirow{2}*{\textbf{Configurations}} & \multicolumn{1}{c}{\textbf{Cat$\rightarrow$Dog}} &  \multicolumn{2}{c}{\textbf{Horse$\rightarrow$Zebra}}\\
			\cmidrule(lr){2-2}\cmidrule(lr){3-4}
			& \textbf{FID} $\downarrow$ & \textbf{FID} $\downarrow$ & \textbf{sec/iter} \\
			
			\midrule
			A $~~$ $(h_7,h_{24})$ & 63.8 &  43.1 & \textbf{0.129} \\ 
			B $~~$ $(h_7, h_{24}), (h_{13}, h_{18})$ & 63.8 &  43.1 & {0.134} \\ 
			\cdashline{1-4}
			C $~~$ EnCo$^\dagger$ & {55.9} & {39.2} & {0.136} \\ 
			\cdashline{1-4}
			D $~~$ uniform ($k=1,\beta=0.0$) & 63.8 &  43.1 & {0.135} \\ 
			\cdashline{1-4}
			E $~~$ default ($k=4,\beta=0.5$) & \textbf{54.7} & \textbf{38.7} & 0.136 \\ 
			
			\bottomrule
		\end{tabular}
	}
	\caption{Ablations on oversampling ratio $k$ and importance sampling ratio $\beta$ of DAG patch sampling strategy.
	}
	\label{tab:abl_main}
\end{table}

EnCo also achieved the highest scores in terms of segmentation indicators on the \emph{Cityscapes}, except for the pixAcc indicator, which is slightly below QS-Attn. This may come from the problem of class imbalance. However, it is worth noting that EnCo is much higher than the other methods in the classAcc metric (37.2 compared to the second place 33.0). This indicates that EnCo uncovers more category-pixel correspondences in the generation. For example, in the last row of Figure~\ref{fig:cmp_main}, EnCo successfully generates the traffic cone represented in yellow in the semantic annotation, while all the other methods fail.

\noindent \textbf{Model efficiency analysis}
We also report the training speed and memory consumption of the different models in Table~\ref{tab:cmp_main}. As can be seen, EnCo exhibits fast training speed with sec/iter 0.14,  which is 2.9$\times$ faster than the reconstruction-based approach of CycleGAN and 1.7$\times$ faster than the Siamese network-based approach of CUT. Meanwhile, the memory usage of EnCo is extremely efficient, with only 2.83 GB compared to the lowest one 2.67 GB. The training efficiency of EnCo comes from its content-constrained design, which requires neither networks for reconstruction nor feature extraction, such as Siamese networks.

\subsection{Ablation Study}

Compared to baselines, our method exhibits superior performance. We design several ablation experiments to analyze each contribution of components in isolation. We mainly conduct ablation experiments on two datasets, \emph{Cat$\rightarrow$Dog} and \emph{Horse$\rightarrow$Zebra}. We report the results of ablation experiments in Table~\ref{tab:abl_main}.

\noindent \textbf{Influence of multi-scale pairwise features constraint.}
We initially investigated the impact of multi-scale pairwise features on generative performance. As shown in rows A-B of Table~\ref{tab:abl_main}, it is evident that the model's performance improves to varying degrees with the inclusion of additional pairs of same-scale features with different semantic levels.

\noindent \textbf{Impact of asymmetric generator architecture.}
We assume the process of encoding and decoding is symmetric rather than the network design, which means that EnCo can be used for asymmetric generators. To prove this, we added a new experiment (EnCo$^{\dagger}$) in Table~\ref{tab:abl_main}, which simulates asymmetry. For compatibility, we use bilinear interpolation and add a simple pre-projection MLP to align a feature pair. Specifically, we selected three non-symmetric feature pairs, $(h_3,h_{24})$, $(h_7,h_{20})$, and $(h_{13},h_{17})$, with $h_3$ having a resolution of 256 and 64 channels, while $h_{24}$ has a resolution and number of channels of 128. As shown in Table~\ref{tab:abl_main}, due to the deliberately asymmetric design, EnCo$^{\dagger}$ exhibits a slight performance degradation compared to EnCo, but it still maintains its robustness and outperforms most of the baselines.

\begin{figure}[t!]
	\begin{center}
		\includegraphics[width=1\linewidth]{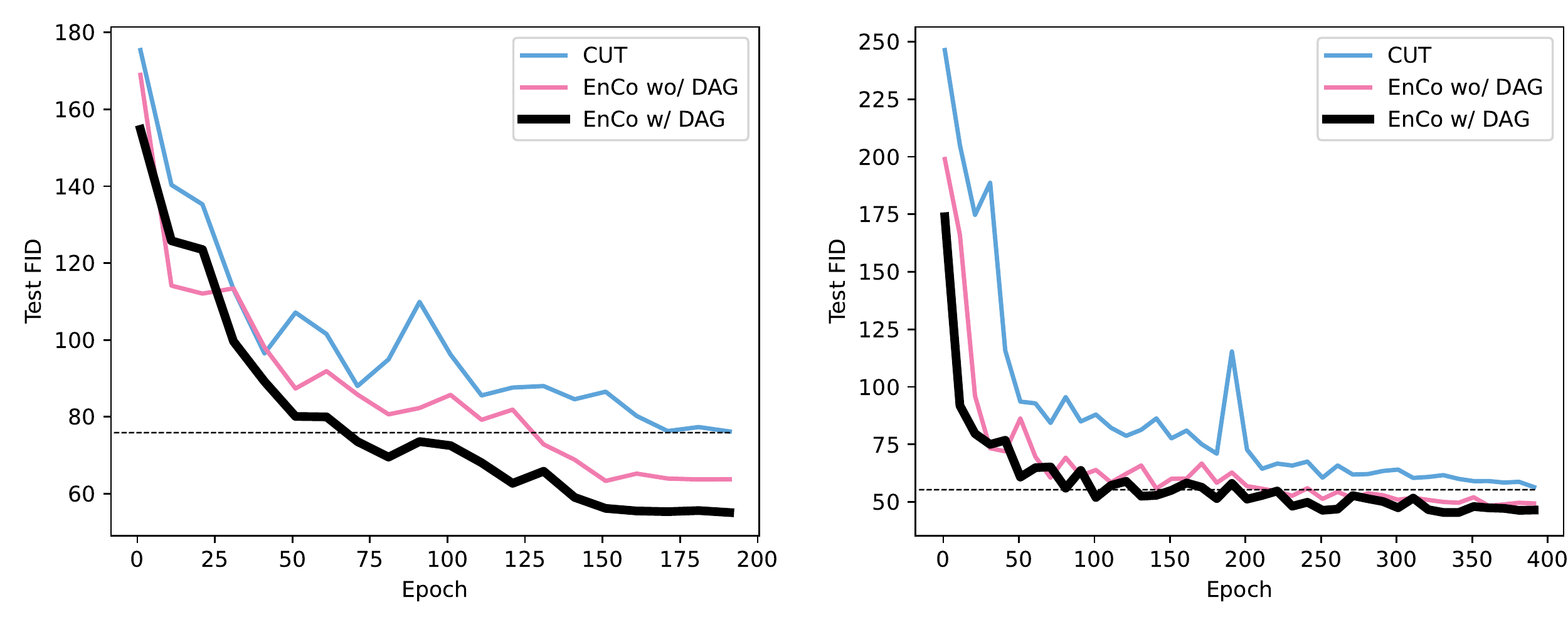}
	\end{center}
	\caption{Comparison of test FIDs over training time in tasks \emph{Cat$\rightarrow$Dog} (left) and \emph{Cityscapes} (right).}
	\label{fig:test_fid}
\end{figure}

\noindent \textbf{Effectiveness of the DAG strategy.}
We also conducted the ablation experiments for the DAG sampling strategy. As can be seen from row D in Table~\ref{tab:abl_main}, when we remove the DAG sampling strategy, \emph{i.e.}, use the random strategy, the performance of the model on each task shows different degrees of degradation compared to the default setting, and training speed was hardly affected, which fully demonstrates the effectiveness of DAG patch sampling strategy. Figure~\ref{fig:vis_sampling} gives an example of how DAG negative sampling strategy affects the generating results. We further visualize in Figure~\ref{fig:test_fid} showing how the test FID changes with training. The results show that without/with DAG, EnCo reached full CUT performance at the 130/65-th (left) and 135/80-th (right) epochs, respectively. This demonstrates well that the DAG strategy accelerates model convergence.

\section{Conclusion}

In this paper, we introduce EnCo, a novel framework for image-to-image tasks. EnCo preseveres content by agreeing on the representational similarity in the latent space of features from the same stage of the encoder and decoder of generator. Compared to current reconstruction-based or Siamese network-based methods, EnCo offers more efficient training, as the constraints are integrated within the generative model. We also rethink the role of the discriminator in the patch sampling and propose a parameter-free discriminative attention-based patch sampling strategy, which incurs almost negligible computational overhead while significantly enhancing the generative performance . Through extensive experiments on multiple popular datasets, we have demonstrated the efficiency and effectiveness of EnCo. We hope EnCo will bring some new thoughts and inspiration to the paradigm of content constrains for unpaired I2I tasks.


\section{Acknowledgments}

This work was supported in part by the National Natural Science Foundation of China under Grant 82073338, and in part by the Sichuan Provincial Science and Technology Department under Grant 2022YFS0384 and 2022YFQ0108. 

\bibliography{EnCo.bib}

\input{appendix}

\end{document}

%% file: appendix.tex
\clearpage
\section{Appendix}

\subsection{More Ablation Study}

In this section, we additionally ablation analyze the effectiveness of some of EnCo's components, including the design of the predictor, the extensibility of EnCo, and the preference of the hyperparameters of the DAG. We mainly conduct ablation experiments on two datasets, \emph{Cat$\rightarrow$Dog} and \emph{Horse$\rightarrow$Zebra}. The results of ablation experiments are reported in Table~\ref{tab:abl_sec} and Table~\ref{tab:abl_samp}.

\noindent \textbf{Design of Predictor.}
When we remove the projection head from EnCo, this means that the features from the encoder and decoder share the same projection head to be projected into the hidden space.However, the removal of the predictor undermines the learning of EnCo (see row A in the Table~\ref{tab:abl_sec}). We believe that the shared projection head cannot decouple the features from the encoder and decoder well. Therefore, EnCo requires an additional nonlinear transformation (i.e., a predictor) that projects the latent features of the decoder into the latent space of the encoder's features. 

\noindent \textbf{Role of Layer Normalization.}
We also try to remove the layer normalization (LN) in the projection and prediction, and as a result, this also harms the generative performance to varying degrees (see row B in the Table~\ref{tab:abl_sec}). We argue that LN plays a role in that it smooths and facilitates feature learning by taking into account the statistical information of other patches in the image.

\begin{figure}[!h]
	\centering
	\includegraphics[width=0.9\linewidth]{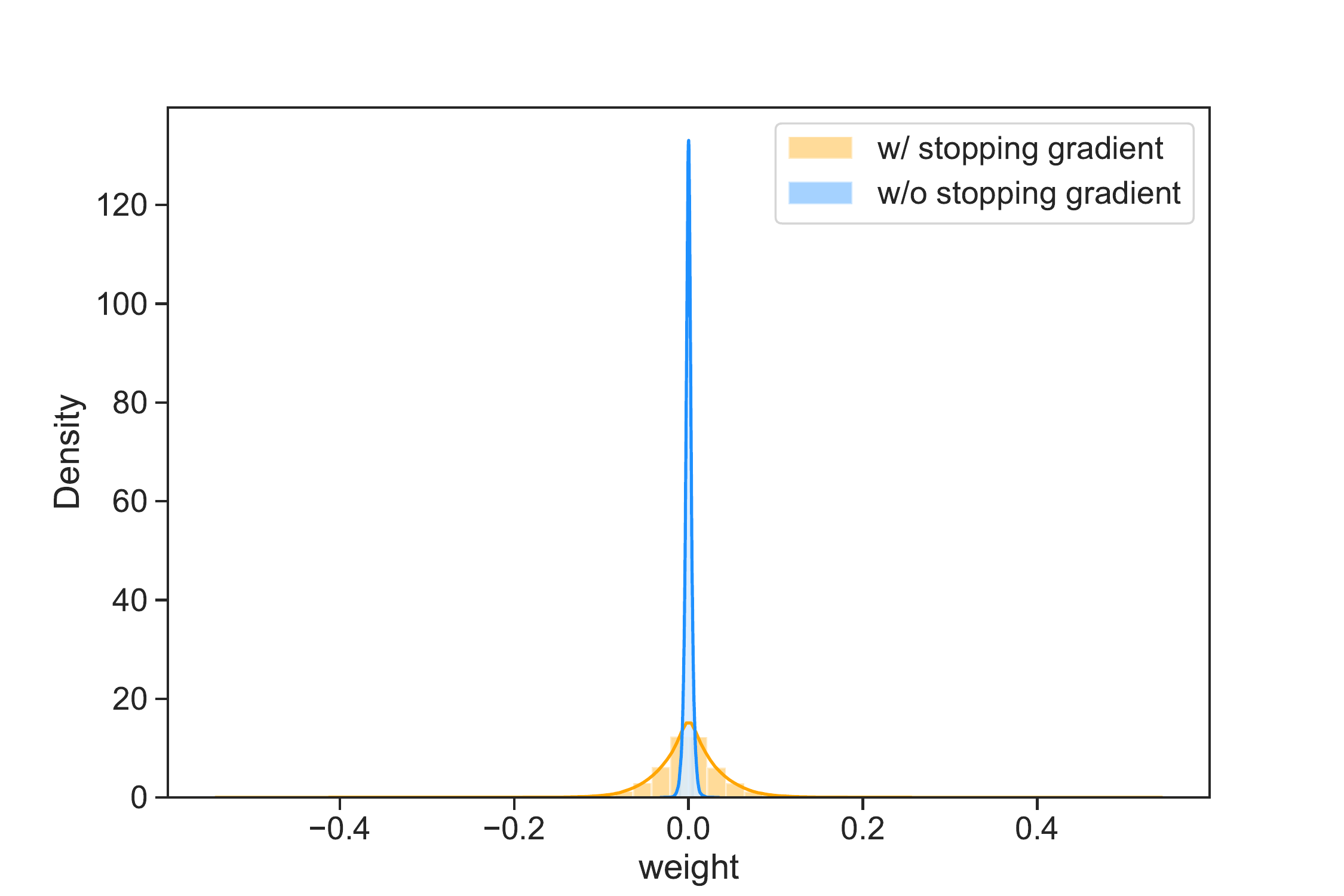}
	\caption{Comparison of density plots.}
	\label{fig:weights}
\end{figure}

\noindent \textbf{Effect of Stopping gradient.}
As we claimed in our paper, SG is a key component in preventing training collapse. Similar ideas are shared in Siamese-based self-supervised learning~\cite{simsiam2020, byol2020}.
To justify this argument, we try not to stop the feature branch from the encoder anymore and find a training collapse occurred (see row C in the Table~\ref{tab:abl_sec}). As a result, stopping the gradient is an important component for successful training of EnCo. Further, we compare the density plot of weights of projections with/without stopping gradient. As shown in Figure~\ref{fig:weights}, projection weights tend to be zeros when gradient stopping is not used, while EnCo avoids trivial solutions.


\noindent \textbf{Extending EnCo with CUT.}
It is easy to combine EnCo to other methods, such as CUT~\cite{park2020cut}. when we replace EnCo's MSE loss function with NCE loss in CUT. At this point, EnCo can be viewed as a lighter version of CUT. we report this result in row D of Table~\ref{tab:abl_sec}. It can be seen that after using EnCo's framework, the CUT's FIDs on the \emph{Cat$\rightarrow$Dog} and \emph{Horse$\rightarrow$Zebra} tasks are significantly reduced ($76.2\to63.8$ and $45.5\to43.1$), while the training speed is greatly reduced also ($0.245\to0.136$). In row E of Table~\ref{tab:abl_sec}, we further added DAG to replace the random sampling strategy in the original CUT, and the model generation performance was further improved. These results amply demonstrate the effectiveness of EnCo and DAG.

\begin{table}[ht]
	\scriptsize
	\centering
	\resizebox{0.93\columnwidth}{!}{
		\begin{tabular}{{l}*{5}{c}}
			\toprule
			\multirow{2}*{Method} & \multicolumn{1}{c}{\textbf{Cat$\rightarrow$Dog}} &  \multicolumn{2}{c}{\textbf{Horse$\rightarrow$Zebra}}\\
			\cmidrule(lr){2-2}\cmidrule(lr){3-5}
			& \textbf{FID} $\downarrow$ & \textbf{FID} $\downarrow$ & \textbf{sec/iter} \\
			
			\midrule
			
			A~~remove predictor & 64.2 &  43.1 & \textbf{0.135} \\ 
			B~~remove layer normalization  & 62.3 &  46.9 & 0.135 \\ 
			C~~w/o stop gradient & - &  - & 0.136 \\ 
			\cdashline{1-5}
			C~~CUT & 76.2 & 45.5 & 0.245 \\
			D~~CUT + EnCo & 63.8 & 43.1 & 0.136 \\
			E~~CUT + EnCo + DAG & 55.5 & 39.1 & 0.136 \\
			
			\cdashline{1-5}
			F~~default & \textbf{54.7} & \textbf{38.7} & 0.136 \\ 
			
			\bottomrule
		\end{tabular}
	}
	\caption{Additional ablations on EnCo components.
	}
	\label{tab:abl_sec}
\end{table}

\begin{table}[h]
	\scriptsize
	\centering
	\resizebox{1\columnwidth}{!}{
		\begin{tabular}{{l}*{5}{c}}
			\toprule
			\multirow{2}*{Method} & \multicolumn{1}{c}{\textbf{Cat$\rightarrow$Dog}} &  \multicolumn{2}{c}{\textbf{Horse$\rightarrow$Zebra}}\\
			\cmidrule(lr){2-2}\cmidrule(lr){3-5}
			& \textbf{FID} $\downarrow$ & \textbf{FID} $\downarrow$ & \textbf{sec/iter} \\
			
			\midrule
			
			uniform ($k=1,\beta=0.0$) & 63.8 &  43.1 & \textbf{0.135} \\ 
			midly biased ($k=4,\beta=0.75$) & 64.9 &  46.7 & 0.136 \\ 
			heavily biased ($k=4,\beta=1.0$) & {68.7} &  45.6 & 0.136 \\ 
			
			\cdashline{1-5}
			small margin ($k=2,\beta=0.5$) & 58.4 & 40.6 & 0.135 \\ 
			large margin ($k=8,\beta=0.5$) & {60.8} & 42.3 & 0.137 \\
			
			\cdashline{1-5}
			default ($k=4,\beta=0.5$) & \textbf{54.7} & \textbf{38.7} & 0.136 \\ 
			
			\bottomrule
		\end{tabular}
	}
	\caption{Ablations on oversampling ratio $k$ and importance sampling ratio $\beta$ of DAG patch sampling strategy.
	}
	\label{tab:abl_samp}
\end{table}

\noindent \textbf{Effectiveness of the DAG strategy.}
In Table~\ref{tab:abl_samp}, we further report the results using different DAG hyperparameters, including the oversampling rate $k$ and importance sampling rate $\beta$. We find that a large importance sampling rate is not friendly to the I2I tasks, which may be a result of over-focusing on hard sample regions and thus the instability of training. Also, we find that the optimized margin yields better performance when the oversampling rate is set to 4.

\begin{table*}[t]
	\scriptsize
	\centering
	\resizebox{0.76\textwidth}{!}{
		\begin{tabular}{{l}*{6}{c}}
			\toprule
			\multicolumn{3}{c}{\textbf{Encoder}} &  \multicolumn{3}{c}{{\textbf{Decoder}}}\\
			\cmidrule(lr){1-3}\cmidrule(lr){4-6}
			\textbf{Name} & \textbf{Operations} & \textbf{Output Size} & \textbf{Name} & \textbf{Operations} & \textbf{Output Size}  \\
			
			\midrule
			
			$h_{0}$ & Input & $256\times 256 \times 3$ & $h_{31}$ & Conv2d,Tanh,Output & $256\times 256 \times 3$  \\ 
			
			$h_{3}$ & Conv2d,IN,ReLU & $256\times 256 \times 64$ & $h_{28}$ &  Conv2d,IN,ReLU & $256\times 256 \times 64$  \\ 
			
			$h_{6}$ & Conv2d,IN,ReLU & $256\times 256 \times 128$ & $h_{25}$ &  Upsample & $256\times 256 \times 128$  \\ 
			
			$h_{7}$ & Downsample & $128\times 128 \times 128$ & $h_{24}$ & Conv2d,IN,ReLU & $128\times 128 \times 128$  \\ 
			
			$h_{10}$ & Conv2d,IN,ReLU & $128\times 128 \times 256$ & $h_{21}$ &  Upsample & $128\times 128 \times 256$  \\ 
			
			$h_{11}$ & Downsample & $64\times 64 \times 256$ & $h_{20}$ & ResnetBlock & $64\times 64 \times 256$  \\ 
			
			$h_{12}$ & ResnetBlock & $64\times 64 \times 256$ & $h_{19}$  & ResnetBlock & $64\times 64 \times 256$  \\ 
			
			$h_{13}$ & ResnetBlock & $64\times 64 \times 256$ & $h_{18}$  & ResnetBlock & $64\times 64 \times 256$  \\ 
			
			$h_{14}$ & ResnetBlock & $64\times 64 \times 256$ & $h_{17}$  & ResnetBlock & $64\times 64 \times 256$  \\ 
			
			$h_{15}$ & ResnetBlock & $64\times 64 \times 256$ & $h_{16}$  & ResnetBlock & $64\times 64 \times 256$  \\ 
			
			\bottomrule
		\end{tabular}
	}
	\caption{{\bf Architecture of the generator.} }
	\label{tab:arch_gen}
\end{table*}

\begin{table}[t]
	\scriptsize
	\centering
	\resizebox{0.42\textwidth}{!}{
		\begin{tabular}{{l}*{3}{c}}
			\toprule
			\textbf{Name} & \textbf{Operations} & \textbf{Output Size}  \\
			
			\midrule
			
			$h_{0}$ & Input & $256\times 256 \times 3$   \\ 
			
			$h_{3}$ & Conv2d,LeakyReLU & $255\times 255 \times 64$  \\ 
			
			$h_{7}$ & Downsample & $128\times 128 \times 64$  \\ 
			
			$h_{6}$ & Conv2d,IN,LeakyReLU & $127\times 127 \times 128$  \\ 
			
			$h_{7}$ & Downsample & $64\times 64 \times 128$  \\ 
			
			$h_{10}$ & Conv2d,IN,LeakyReLU & $63\times 63 \times 256$  \\ 
			
			$h_{11}$ & Downsample & $32\times 32 \times 256$  \\ 
			
			$h_{10}$ & Conv2d,IN,LeakyReLU & $31\times 31 \times 256$  \\ 
			
			$h_{12}$ & Conv2d & $30\times 30 \times 1$ \\  
			
			\bottomrule
		\end{tabular}
	}
	\caption{{\bf Architecture of the discriminator.}}
	\label{tab:arch_dis}
\end{table}

\subsection*{Architecture of Networks\label{appendix:arch}}

Following CycleGAN and CUT, We use a ResNet-like generator. The detailed  generator architecture is shown in Table~\ref{tab:arch_gen}. Conv2d uses kernel size 3, padding 1 and stride 1 expect for the first Conv2d using a large kernel size 7, paddding 3 and stride 1. IN means instance normalization. ResnetBlock consists of Conv2d, IN, ReLU, Conv2d and IN. We decompose the generator $G$ into two parts, the encoder and decoder. CUT assigns the first 5 ResnetBlocks to the encoder while we assign the 5-th ResnetBlock to the decoder for symmetry reason. We also apply a PatchGAN discriminator as CycleGAN and Pix2Pix. As  shown in Table~\ref{tab:arch_dis}, the discriminator in our model contains three down-sampling blocks and outputs a $30\times30$ feature map, each element in which correspond to a $70\times70$ patch in the input image.

\subsection*{Evaluation Details}

We list the details of our evaluation protocol. Fr\'{e}chet Inception Distance (FID~\cite{heusel2017ttur}) is computed by measuring the mean and variance distance of the generated and real images in a deep feature space. We first resize the images to $299\times$299, and then feed them into a pretrained Inception model to extract deep features. Here, we used the default setting of \url{https://github.com/mseitzer/pytorch-fid} to compute the FID score on test set images. 

For semantic segmentation metrics (\emph{i.e.}, mAP, pixAcc and classAcc) on the \emph{Cityscapes} dataset, we used a pretained DRN-D-22 model to segment the generated images and compared with the corresponding ground truth labels. Specifically,  the pretained DRN-D-22 model are provided in \url{https://github.com/fyu/drn}, which was trained with batch size 32 and learning rate 0.01, for 250 epochs at 256x128 resolution. Before passed to the segmentation model, we resized the input images to 256x128 using bicubic downsampling and the ground truth labels were downsampled to the same size using nearest-neighbor sampling.

\begin{algorithm}[!ht]
	\caption{DAG Pseudocode, PyTorch-like}
	\label{alg:DAG}
	\definecolor{codeblue}{rgb}{0.25,0.5,0.5}
	\definecolor{codekw}{rgb}{0.85, 0.18, 0.50}
	\lstset{
		backgroundcolor=\color{white},
		basicstyle=\fontsize{7.5pt}{7.5pt}\ttfamily\selectfont,
		columns=fullflexible,
		breaklines=true,
		captionpos=b,
		commentstyle=\fontsize{7.5pt}{7.5pt}\color{codeblue},
		keywordstyle=\fontsize{7.5pt}{7.5pt}\color{codekw},
	}
	\begin{lstlisting}[language=python]
		# H: height, W: width, C: dimension
		# K: number of patches to sample
		# k: oversampling ratio
		# beta: importance sampling ratio
		# feat: input tensor (H, W, C)
		# score_map: spatial attention tensor, (H, W)
		
		H, W, C = feat.shape
		feat_reshape = feat.flatten(0, 1)
		# oversampling
		points_os = torch.randperm(H * W)[:k * K]  
		scores_os = score_map.flatten()[points_os]
		# ranking
		_, indices = torch.sort(scores_os, descending=False)
		# importance sampling
		idx_DAG = points_os[indices[:beta * K]]
		# covering
		idx_random = torch.randperm(H * W)[(beta - 1) * K:]
		idx_sampled = torch.cat([idx_DAG, idx_random])
		
		feat_to_return = feat_reshape[idx_sampled, :]
		
	\end{lstlisting}
\end{algorithm}

\subsection*{Discriminative Attention-guided Patch Sampling Algorithm}
We provide the pseudo-code of the DAG patch sampling strategy in PyTorch style in Algorithm~\ref{alg:DAG}.

\subsection*{More Generated Results}

\begin{figure*}[ht]
	\centering
	\includegraphics[width=\linewidth]{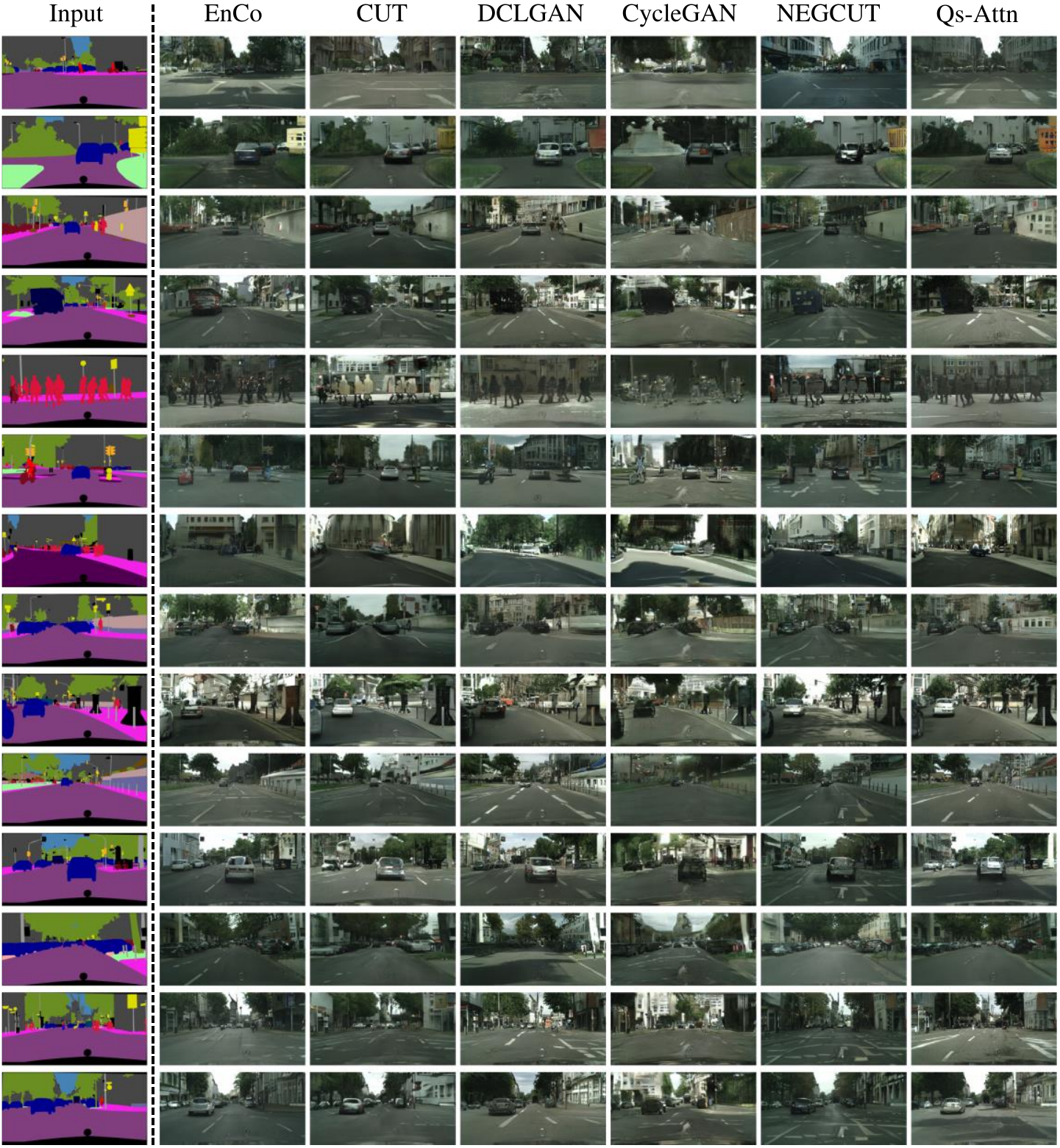}
	\caption{{\bf More results of qualitative comparison on the \emph{Cityscapes} dataset.}
	}
	\vspace{-4mm}
	\label{fig:appendix_city}
\end{figure*}

\begin{figure*}[ht]
	\centering
	\includegraphics[width=\linewidth]{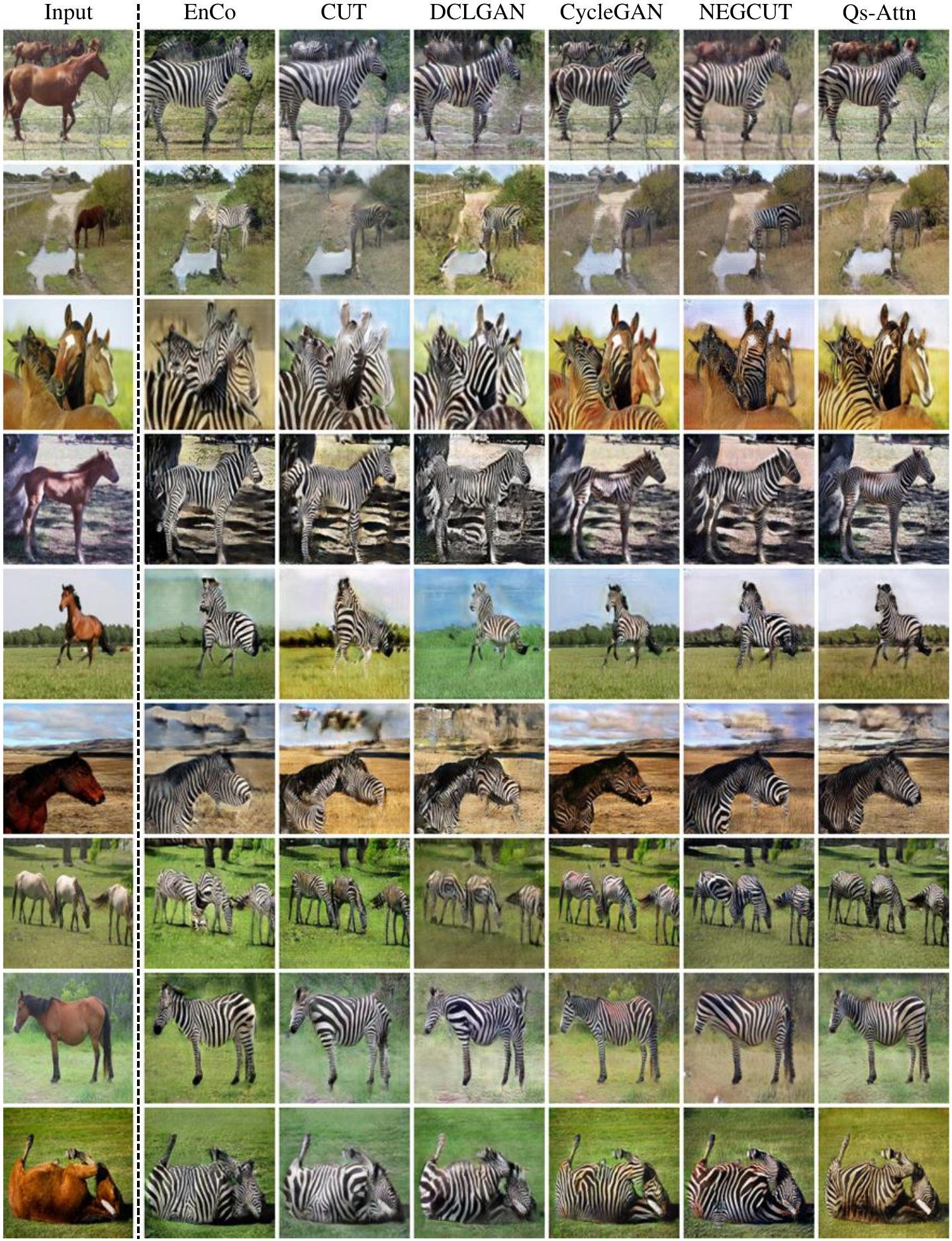}
	\caption{{\bf More results of qualitative comparison on the \emph{Horse$\rightarrow$Zebra} dataset.}
	}
	\vspace{-4mm}
	\label{fig:appendix_horse}
\end{figure*}

\begin{figure*}[ht]
	\centering
	\includegraphics[width=\linewidth]{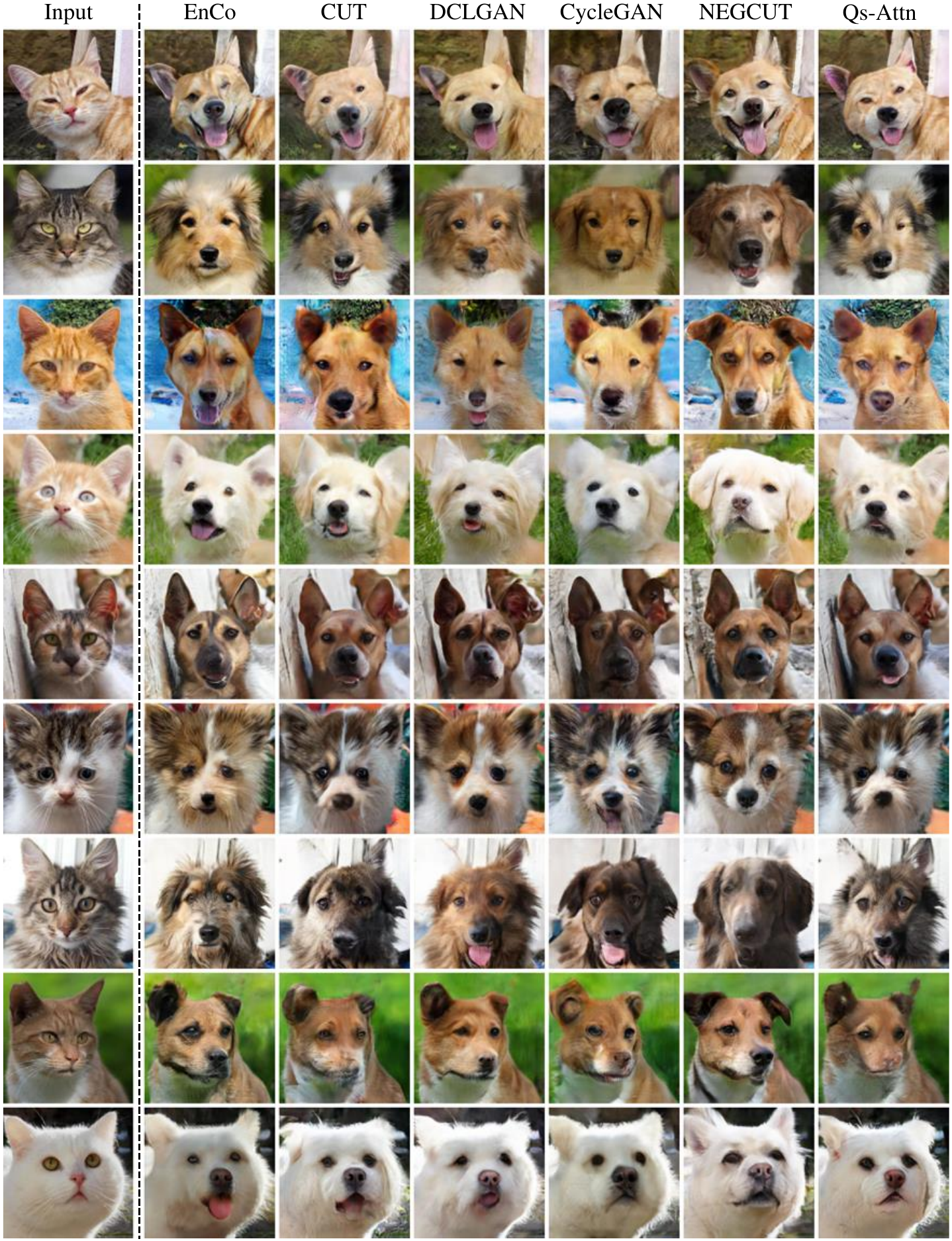}
	\caption{{\bf More results of qualitative comparison on the \emph{Cat$\rightarrow$Dog} dataset.} 
	}
	\vspace{-4mm}
	\label{fig:appendix_cat}
\end{figure*}

We show additional comparison results for \emph{Cityscapes}, \emph{Horse$\rightarrow$Zebra} and \emph{Cat$\rightarrow$Dog} tasks in Figure~\ref{fig:appendix_city}, Figure~\ref{fig:appendix_horse} and Figure~\ref{fig:appendix_cat}, respectively. This is an extension of Figure~5 in the main paper. We only show the results of EnCo, CUT, DCLGAN, FSeSim and QS-Attn in these figures. They achieved challenging results in Table 1 in the main paper.

\clearpage

\clearpage

\clearpage